%% file: arxiv.tex
\documentclass[10pt,twocolumn,letterpaper]{article}

\usepackage{cvpr}
\usepackage{times}
\usepackage{epsfig}
\usepackage{graphicx}
\usepackage{amsmath}
\usepackage{amssymb}

\usepackage{multirow}
\usepackage{booktabs}

\usepackage{caption}

\usepackage{xcolor}

\DeclareMathOperator*{\argmax}{argmax}
\DeclareMathOperator*{\argmin}{argmin}

\newcommand{\val}{\texttt{val}\xspace}
\newcommand{\test}{\texttt{test}\xspace}

\newcommand{\aug}{\texttt{train-aug}\xspace}

\usepackage[pagebackref=true,breaklinks=true,letterpaper=true,colorlinks,bookmarks=false]{hyperref}

 \cvprfinalcopy 

\def\cvprPaperID{2246} 

\ifcvprfinal\fi
\begin{document}

\title{Diverse Sampling for Self-Supervised Learning of
    Semantic Segmentation}

\author{Mohammadreza Mostajabi \thanks{Authors contributed equally} \and Nicholas Kolkin \footnotemark[1] \and Gregory Shakhnarovich\\ \and
Toyota Technological Institute at Chicago\\
{\tt\small \{mostajabi, nick.kolkin, greg\}@ttic.edu}
}

\maketitle

\begin{abstract}
We propose an approach for learning category-level semantic
segmentation purely from image-level classification tags indicating
presence of categories. It exploits localization
cues that emerge from training classification-tasked convolutional
networks, to drive a ``self-supervision'' process that automatically
labels a sparse, diverse training set of points likely to belong to classes of
interest. Our approach has almost no hyperparameters, is modular, and
allows for very fast training of segmentation in less than 3 minutes. It obtains competitive
results on the VOC 2012 segmentation benchmark. More, significantly the modularity and fast training of our framework allows new classes to efficiently added for inference.  
\end{abstract}

\input{Intro}

\input{Related}

\input{Methods}

\input{Results}

\input{conclusions}

\input{Acknowledgments}

\clearpage
\bibliographystyle{abbrv}
\bibliography{weakseg}

\end{document}

%% file: Intro.tex
\section{Introduction}\label{sec:intro}

The problem of semantic segmentation (category-level labeling of
pixels) has attracted significant attention. Most recent progress
can be attributed to advances in
deep learning and to availability of large, manually
labeled data sets. However, the cost and complexity
of annotating segmentation are significantly higher than that for
classification; consequently, we have orders of magnitude more images
and categories in classification data sets such as ImageNet~\cite{russakovsky2014imagenet} or Places2~\cite{zhou2014learning}, than in
segmentation data sets, such as VOC~\cite{everingham2015pascal} or
MS-COCO~\cite{lin2014microsoft}.

Given this gap, and the objective difficulty in rapidly closing it,
many researchers have considered weakly supervised
segmentation, where the goal is still pixel-level labeling at test
time, but only spatially coarses annotations are available at training
time. Common examples of such annotations include partial
annotations, in which only a subset of pixels is labeled; bounding
boxes, where a square with an associated label is drawn around objects
of interest; and image tags, where labels provide no spatial information
and simply indicate whether or not a particular class is present
\emph{somewhere} in the image. We focus on this last, arguably
weakest, level of per-image supervision.

There is mounting evidence that this task, while difficult, is not hopeless. 
Units sensitive to object localization have been shown to emerge as part of the
representations learned by convolutional neural networks (CNNs) trained for
image classification~\cite{zhou2015object}. Furthermore, some
localization methods demonstrate the utility of features learned by
classification CNNs by using them to achieve competitive results~\cite{oquab2015object,zhou2015object}.

Our method is inspired by recent work~\cite{russakovsky2015what} 
demonstrating that reasonable segmenation accuracy could
be achieved with very few point-wise labels provided by human annotators,. In this paper we propose an automatic version of this
idea, replacing human annotators with an automatic labeling
procedure. Our approach starts by learning noisy localization
networks separately for each foreground class, trained solely with image-level classification tags, and using a novel multiple
instance learning loss (global softmax, Section~\ref{sec:method}) adapted for the segmentation task. By combining the localization evidence provided by
 these networks with a novel diversity sampling procedure, we obtain
 sparse, informative, and accurate set of labeled pixels. We can use
 these samples to
 rapidly train a fully-convolutional multi-class pixel-wise label predictor
 operating on hypercolumn/zoomout representation of image
 features~\cite{hariharan2015hypercolumns,mostajabi2015feedforward} in less than 3 minutes. 

In contrast to much previous work, our approach is simple and
modular. It almost
entirely lacks hyper-parameters like thresholds and weighting
coeficients. It also allows for easy addition of new
foreground classes or incorporation of more image examples for some classes,
without the need to retrain the entire system. We also avoid complex
integration with externally trained components, other than the basic
ImageNet-trained neural network we use to extract pixel
features. Consequently while competitive models take hours to train, our framework takes less than 3 minutes. Despite this simplicity we obtain results on VOC 2012 data
set that improve upon
most of previous work on image-level weak supervision of
segmentation. 

\input{Fig_Sel_Results}

%% file: Fig_Sel_Results.tex
\begin{figure*}[!th]
  \begin{minipage}[c]{.75\linewidth}
  \begin{minipage}[c]{.23\linewidth}
    \includegraphics[width=.95\linewidth]{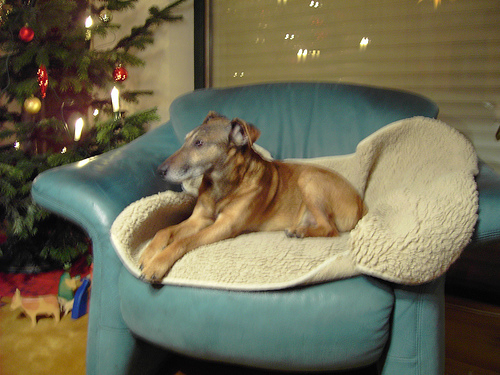} 
  \end{minipage}%
  \begin{minipage}[c]{.5\linewidth}
    \includegraphics[width=.99\linewidth]{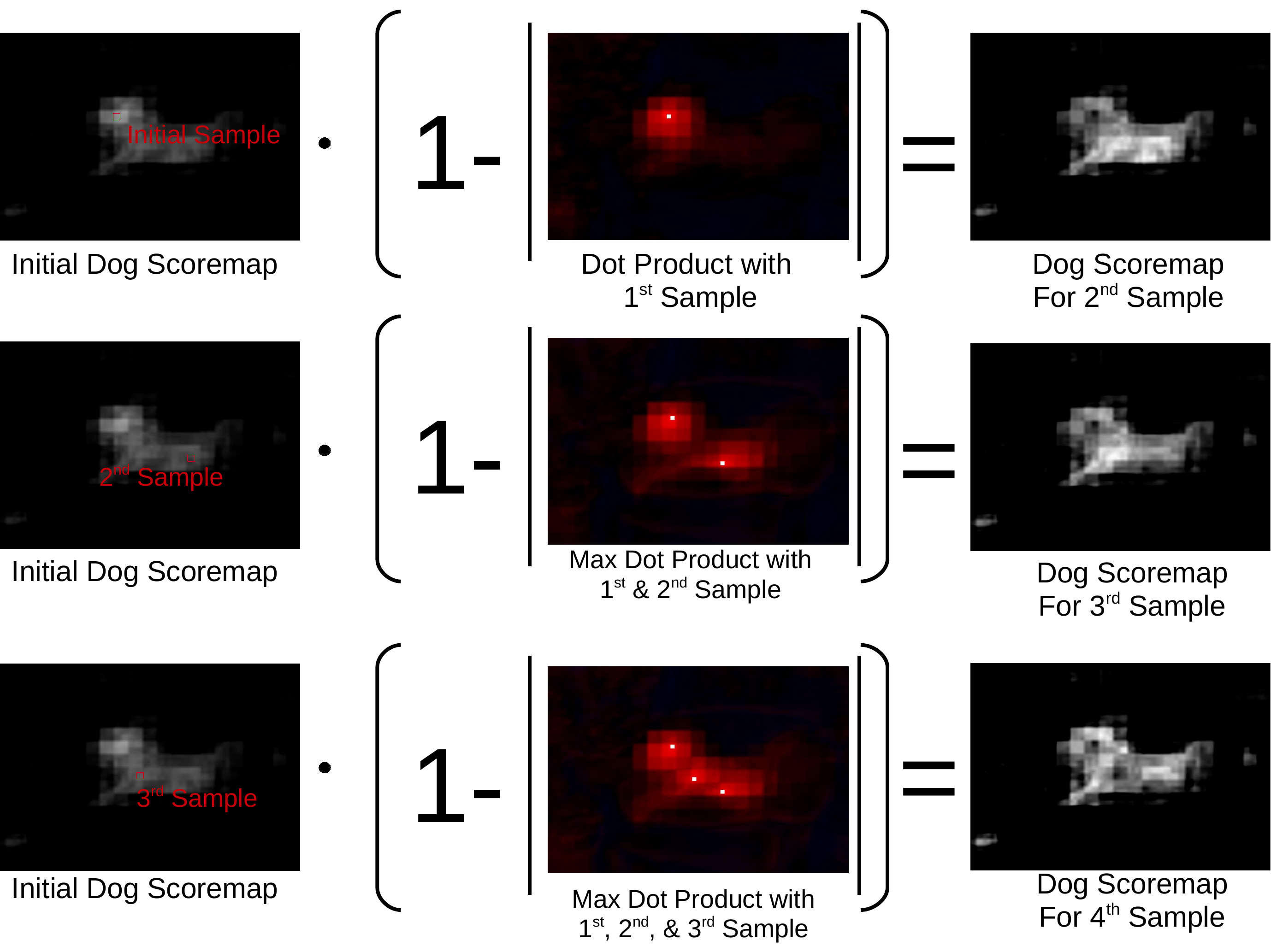}
  \end{minipage}%
  \begin{minipage}[c]{.23\linewidth}
\centering
    \includegraphics[width=.95\linewidth]{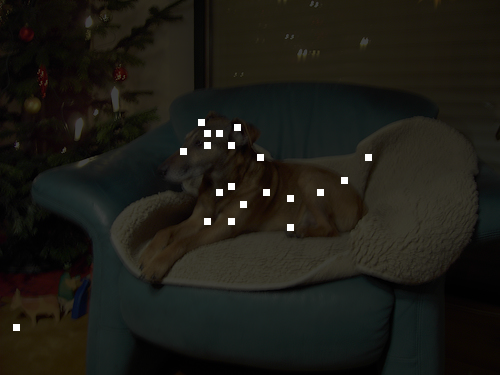} 
  \end{minipage}    
  \end{minipage}%
  \begin{minipage}[c]{.25\linewidth}
\caption{Our diverse sampling procedure. Left: input image, labeled as containing class
  \texttt{dog}. Middle: first four steps of diverse sampling procedure
  for \texttt{dog}, starting with the map
  $S(\cdot,\text{\texttt{dog}})$. Right: sample of 20 diverse
points.}
\label{fig:divsample}
    
  \end{minipage}

\end{figure*}

%% file: Related.tex
\section{Background}\label{sec:related}
Semantic segmentation has seen major progress in recent years,
at least as measured by benchmark performance, nearly doubling the
accuracy from 2012~\cite{boix_ijcv12,carreira2012semantic} to today's leading
methods~\cite{long2015fully,chen2015semantic,zheng2015conditional}. Much
of this progress can be attributed to
(re)introduction of convolutional neural networks. The availability of
training data with manually annotated segmentation
masks, in particular VOC\cite{everingham2015pascal} and recently MS-COCO~\cite{lin2014microsoft}, has been instrumental in these developments, however
recent work has shown that training on weaker annotations, such as
partial labeling~\cite{russakovsky2015what}, object bounding boxes~\cite{papandreou2015weakly,dai2015boxsup,khoreva2016weakly} or other
cues such as object sizes~\cite{pathak2015constrained}, can
produce results very close to those using strongly-supervised training.
However, closing this gap with strongly-supervised methods trained exclusively on image-level
tags, which is the regime we consider in this paper, remains more
challenging.

Our work was in part inspired by the experiments
in~\cite{russakovsky2015what} showing that very sparse point-wise
supervision allows training reasonable CNN-based segmentation
models. Our approach aims to replace manual annotation with
``self-supervision'' obtained automatically from image-level tags. Similarly to other recent work, we obtain
self-supervision by leveraging the recently established observation:
CNNs trained for image classification tasks appear to contain internal
representations tuned to object
localization~\cite{bergamo2014self,zhou2015object,oquab2015object,cinbis2015weakly}. These
representations have been Combined
with a pooling strategy to obtain self-supervision, which can be used
to train a segmentation model, 
often with a variant of
the Expectation-Maximization
algorithm~\cite{papandreou2015weakly,xu2015learning} or with multiple instance learning
(MIL)~\cite{pathak2014fully,pinheiro2015image,kim2016deconvolutional,kolesnikov2016seed}.

Some recent methods combine image-level supervision with additional
components, such as object proposals~\cite{pinheiro2015image}, saliency~\cite{wei2015stc} or objectness
measures~\cite{russakovsky2015what}. Most of these components require
localization annotations, such as bounding boxes and/or segmentation
masks, which introduces a dependency on additional, often expensive
annotations beyond image-level tags.
In contrast, our approach is simple and modular, and does not require any external
systems, other than the initial CNN~\cite{simonyan2014very} pretrained on the ImageNet
classification task~\cite{russakovsky2014imagenet}, making the
entire pipeline independent of any requirements beyond image-level annotations.

The most established
benchmark for this task remains the VOC 2012 data set, with the
standard performance measure being intersection over union averaged
over 21 classes (mIoU),
\footnote{MS-COCO
is larger in categories and images, but at the moment does not allow
for a proper category-level semantic segmentation evaluation, due to its focus on
instance-level detection.} which can be reported on \val or on \test
set. The latter arguably more rigorous since the labels are withheld
and evaluation frequency limited. In
Section~\ref{sec:exp} we show that despite its simplicity and
efficiency, our approach
outperforms most competing methods on this benchmark.

Finally, there is a body of work related to exploring and exploiting
diversity in learning and vision. Most relevant to our work is
the DivMBest algorithm~\cite{batra12diverse,yadollahpour2013discriminative} which somewhat resembles our procedure
for greedy diverse sampling of points described in
Section~\ref{sec:sampling}. The form of the diversity-infused
objective and the context are quite different, however: DivMBest is used to sample assignments in a
structured prediction settings for a single input example, whereas we
aim to sample examples (image points); in applications of DivMBest the
diverse samples are typically fed to a post-processing stage like
reranking whereas in our case, the diverse sample is directly used as
a training set for a segmentation algorithm.

%% file: Methods.tex
\section{Automatic pointwise self-supervision for segmentation}\label{sec:method}
The basis of our self-supervision method is the localization maps
obtained for each of the foreground classes. These maps are sampled
for each class, and the resulting sparse set of
point-wise labels on the training images is used to train the final
segmentation network. We describe these steps in detail below.

\subsection{Learning localization with image-level tags}\label{sec:localization}
 We start by extracting an
image feature map using a pre-trained fully convolutional
network. Then for each foreground class $c$, we 
construct a per-location localization network on top of these features, which
outputs two scores per location $i$: $S(c,i)$ for foreground, and
$\bar{S}(c,i)$ for background (which in this case means anything
other than the foreground class at hand). 

An obvious next step now is
to convert these scores into image-level foreground probability, using
some sort of pooling scheme; this can then be used to compute
image-level classification log-loss and backpropagate it to the
localization network. We consider two such schemes.

\setlength{\parskip}{0cm}

\paragraph{The per-pixel softmax model} We can convert the scores into
per-location posterior probabilities using the standard
(over-parameterized) softmax model, and apply max pooling over the
resulting probability map:
\begin{equation}
  \label{eq:softmax}
p(c)\,=\,\max_i\frac{\exp S(c,i)}{\exp S(c,i)\,+\,\exp \bar{S}(c,i)}.
\end{equation}
This can be interpreted as requiring that for images containing
foreground, the network assign at least one location high probability
while for images without the foreground, all locations must have low
probability. The background scores $\bar{S}$ do not have a direct
meaning other than to normalize the probabilities. 

\paragraph{The global softmax model} An alternative is to apply
max-pooling \emph{separately} for the two score maps, and convert the
maxima to the image-level probability:
\begin{equation}
  \label{eq:globalsoftmax}
  p(c)\,=\,\frac{\max_i \exp S(c,i)}{\max_j \exp S(c,j)\,+\,\max_l\exp
    \bar{S}(c,l)}.
\end{equation}
This model no longer is equivalent to the per-location softmax, and in
fact does not provide per-location probability map. It specifically
encourages the background scores to be high for images without the
foreground. It also routes the gradient of the loss via two locations
in each image, instead of one with~\eqref{eq:softmax}, and therefore may facilitate faster training.

It is worth noting that this approach does not include an explicit
``background localization'' model. Background is defined separately
for each foreground class as its complement, and jointly as the
complement of all foreground classes. Adding another foreground class
would require only training one new localization model for that class;
the definition of segmentation background would then automatically be
updated, and reflected in the sampling process described below.


\input{diverseSampling}

\subsection{Training semantic segmentation using self-supervision}
Once the self-supervision labels are obtained, we are ready to train
the segmentation as a per-pixel fully convolutional multi-class
(including background) classifier
network, with receptive field of 1$\times$1. This can be done by using
the standard convnet training machinery, with zero-masking applied to
the loss in locations where no labels are available; whether to
fine-tune the underlying network that extracts the visual features per
location is a choice.

%% file: diverseSampling.tex
\subsection{Sampling strategies for self-supervision}\label{sec:sampling}
We now consider the goal of translating class-specific score maps to
supervisory signal for semantic segmentation. Our general framework
for this will be to select a sparse set of locations from the training
images, for which we will assign class labels. The segmentation
predictor is
then trained by learning to map the image features for these
locations to the assigned class labels.

Let $S(i,c)$ be the score at spatial index $i$
for a class $c$ produced by our image-level classification
model. One approach would be to densely label $y_i=
\argmax_{c}S(i,c)$. Background requires a separate treatment: it is
not a ``real'' class, rather it's defined by not being one of the
foreground classes. Hence, we do not have a separate model for it, and
instead can assign background labels to pixels in which no foreground
class attains a sufficient score: $y_i=bg$ if $\max_cS(i,c)<\tau$. 
This simplistic strategy has two problems: (1)
 some classes may have systemically lower scores than others, and
(2) it is unclear how to optimally set the value of $\tau$.

However, our hypothesis is that while the scoremaps provide only coarse
localization, and an inconsistent level of confidence across images
and classes, the maximum activations of a class scoremap when that
class is present appear to reliably correspond with pixels containing
the correct class. (We verified this qualitatively, on a few classes and
a number of
training images). So, an alternative approach is to label as $c$ the
$k$ locations
corresponding with highest scores for class $c$. However the size of objects varies widely
across images, and it isn't clear what $k$ should be. If $k$ is too
high, the labels will be very noisy. If $k$
is too low,  most of the pixels will be tightly clustered 
portions of each class, e.g.,  wheels of cars, or
 faces of people; training on such examples is much less effective
because many of the samples will highly correlated. 

The method we propose here alleviates these problems by relying on
\emph{diversity sampling}. Let $\mathbf{z}_i$ be the image feature vector at spatial
index $i$, normalized to unit norm, and let $F$ be the set of foreground classes present in the
image. For each class $c\in F$ we define
the  $k$th sampled location $x^c_k$ from that image by induction:
\begin{eqnarray}\label{eq:div-fg}
  x^c_1\,&=\,\argmax_i S(i,c),\\
  x^c_k\,&=\,\argmax_i\,\left\{S(i,c)\left[1-\max_{k'<k}
  \left| \mathbf{z}_i\cdot\mathbf{z}_{x^c_{k'}} \right|\right]\,\right\}.
\end{eqnarray}
where $\cdot$ denotes the dot product.
In other words, we aim to select points with the highest scores for
class $c$, but penalize them for similarity to
previously selected points for that class. This encourages selection of a set of examples
that jointly provides more information (due to diversity), and is likely
to be accurate (due to high scores). 

The definition of similarity as
dot product in feature space, rather than as spatial
proximity, is important. If features at two points 
are similar, they will appear as similar to the eventual pixel classifier
(since it operates on those features), regardless of how close they
are in the image; labeling only one
of them is sufficient. On the other hand, two points nearby may have
very different feature representations and thus it is beneficial to label both to give the pixel
classifier more information.

This approach also naturally leads to a threshold-free method for
selecting background points. We let the $k^{th}$ sample, $x_{k,bg}$
for be defined as:
\begin{equation}
  \label{eq:div-bg}
  x^{bg}_k\,=\,\argmin_i \max\left\{\,
    \max_{c\in F,k'}\left|\mathbf{z}_i\cdot\mathbf{z}_{x^c_{k'}}\right|\,,\,
    \max_{k''<k}\left|\mathbf{z}_i\cdot\mathbf{z}_{x^{bg}_{k''}}\right|
    \right\}
\end{equation}
where $k'$ ranges over indices of selected points for each foreground
class, and $k''$ ranges over points selected so far for the
background. This simply searches at each step for the image location
most dissimilar to any foreground points -- thus maximizing the chance
of correctly identifying background points, and to any other
background points selected -- thus maximizing diversity. Our diversity
sampling strategy is illustrated in Figures~\ref{fig:divsample},~\ref{fig:sampling}.

%% file: Results.tex
\begin{figure*}[!th]
  \begin{minipage}[c]{.75\linewidth}
    
  \centering
  \begin{tabular}{cccccc}
\includegraphics[width=.15\textwidth]{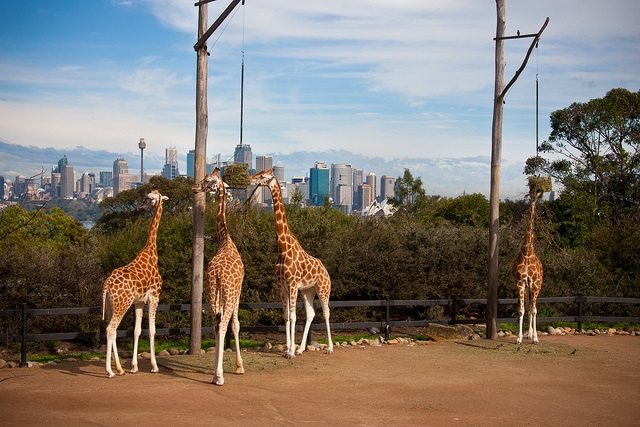} &
 \includegraphics[width=.15\textwidth]{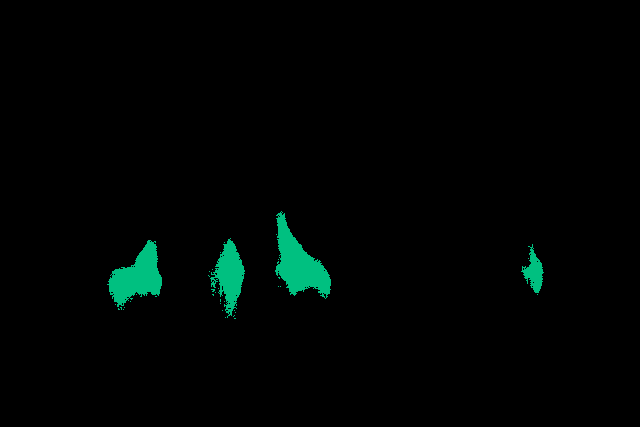}&
\includegraphics[width=.15\textwidth]{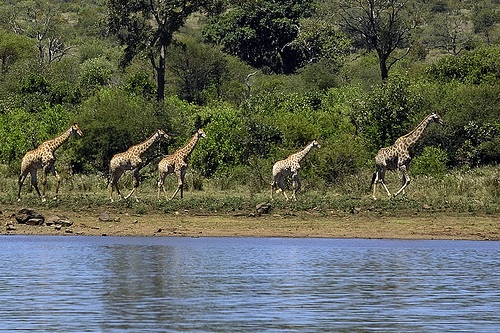} &
 \includegraphics[width=.15\textwidth]{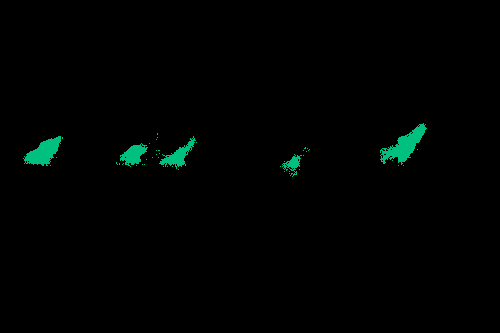}\\
\includegraphics[width=.15\textwidth]{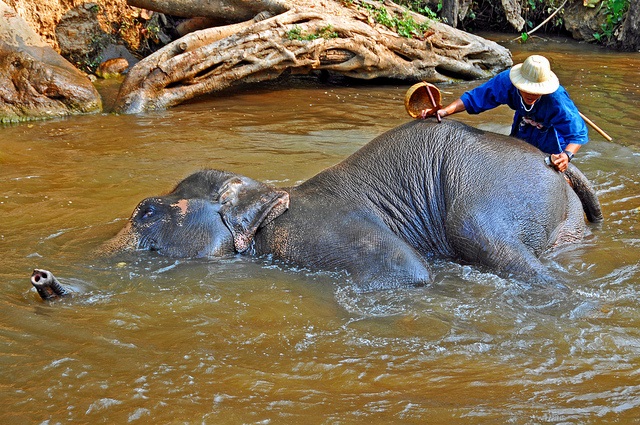} &
 \includegraphics[width=.15\textwidth]{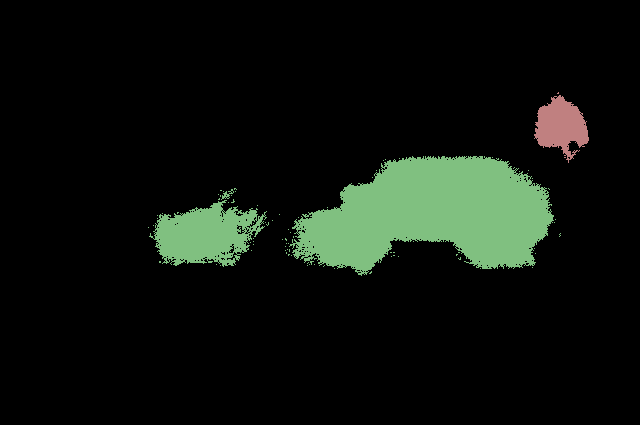}&
\includegraphics[width=.15\textwidth]{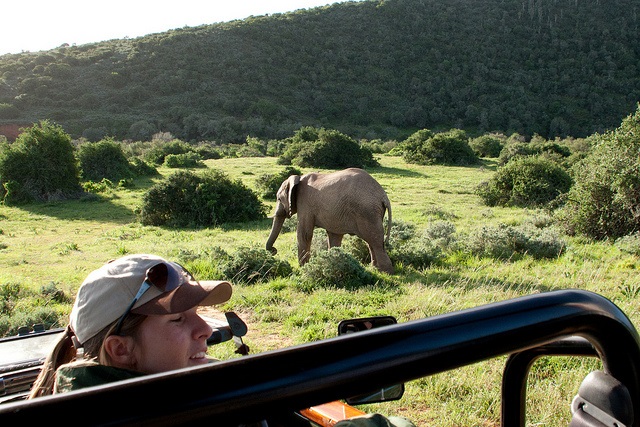} &
 \includegraphics[width=.15\textwidth]{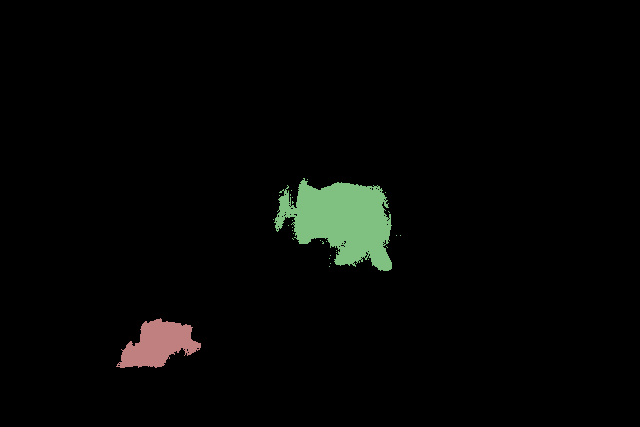}\\
\includegraphics[width=.15\textwidth]{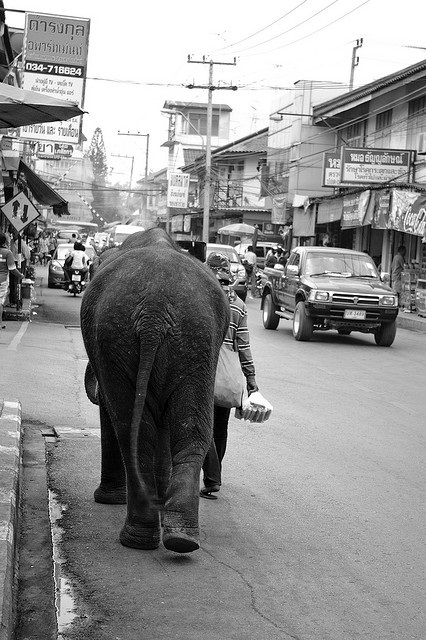} &
 \includegraphics[width=.15\textwidth]{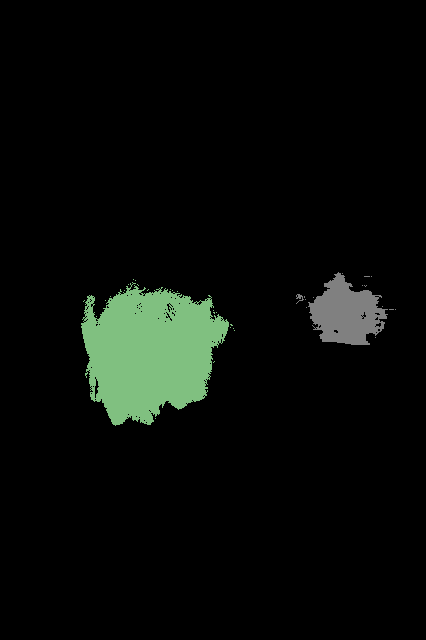}&
\includegraphics[width=.15\textwidth]{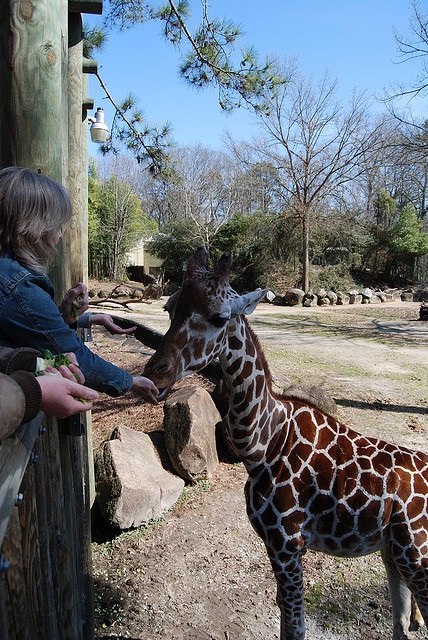} &
 \includegraphics[width=.15\textwidth]{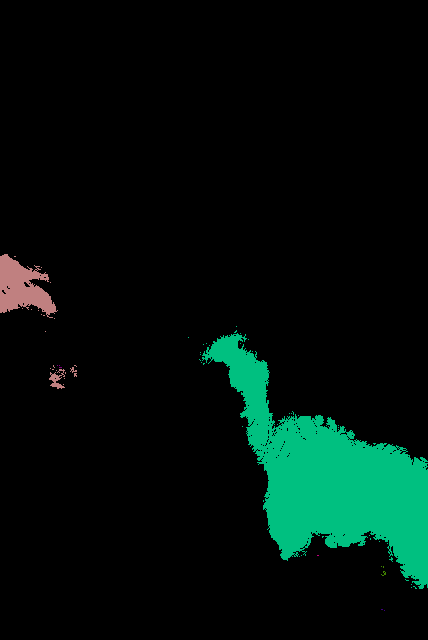}
  \end{tabular}
  \captionof{figure}{Example segmentations on extra classes from MS-COCO dataset added to the segmentation model trained on VOC dataset.}
  \label{fig:extraseg}
  \end{minipage}%
  \begin{minipage}[c]{.25\linewidth}
    
 \begin{tabular}{|l|l|}
\hline
    Model & mIoU\\
\hline
Pixel softmax & 38.0 \\
Global softmax & 40.6\\
\hline
  \end{tabular}
  \captionof{table}{Comparison of localization models on VOC 2012 \val}
  \label{tab:locmodels}

  \begin{tabular}{|l|l|}
\hline
    Sampling & mIoU\\
\hline
Dense & 15.0\\
Spatial & 33.4\\
Top $k=20$ & 30.7\\
Diverse, $k=20$& 40.6\\
\hline
  \end{tabular}
  \captionof{table}{Comparison of point sampling strategies for self-supervision}
  \label{tab:sampling}

\end{minipage}
\end{figure*}

\begin{table}[!th]

\setlength\tabcolsep{3pt}
  \begin{tabular}{|l|l|l|l|l|l|}
\hline
    $k$&1 & 5&10&20 & 50\\\hline
    mIoU & 35.1& 37.2&39.3  &  40.6 & 40.4\\
\hline
  \end{tabular}
    \captionof{table}{mIoU on VOC 2012 \val as a function of $k$ points in
      diverse sampling, with global softmax model}
    \label{tab:miou-vs-k}

\end{table}

\newcommand{\setupSampIm}[3]{ input & #2 & #3 &  background\\  \includegraphics[width=.22\textwidth]{Figures/SampIms/#1}&  \includegraphics[width=.22\textwidth]{Figures/SampIms/#1_#2_inMap}&  \includegraphics[width=.22\textwidth]{Figures/SampIms/#1_#3_inMap}&   \includegraphics[width=.22\textwidth]{Figures/SampIms/#1_background_inMap}\\   &\includegraphics[width=.22\textwidth]{Figures/SampIms/#1_#2_output}&  \includegraphics[width=.22\textwidth]{Figures/SampIms/#1_#3_output}&  \includegraphics[width=.22\textwidth]{Figures/SampIms/#1_background_output}\\ }

\begin{figure*}[!th]
  \centering
  \begin{tabular}{cccc}
\setupSampIm{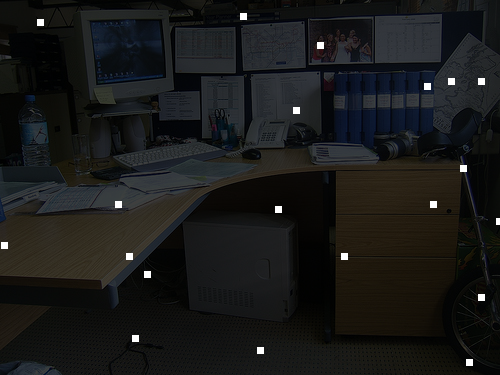}{bottle}{tvmonitor}
\setupSampIm{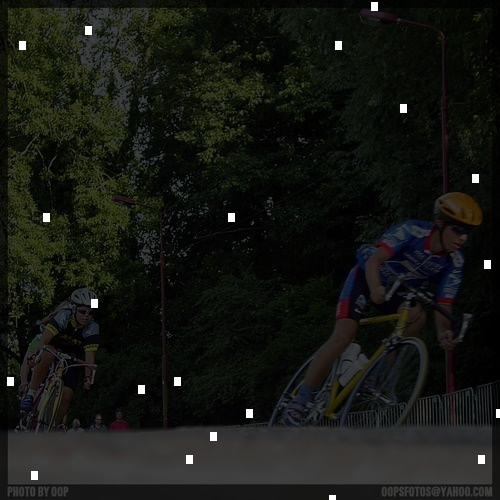}{person}{bicycle}
\setupSampIm{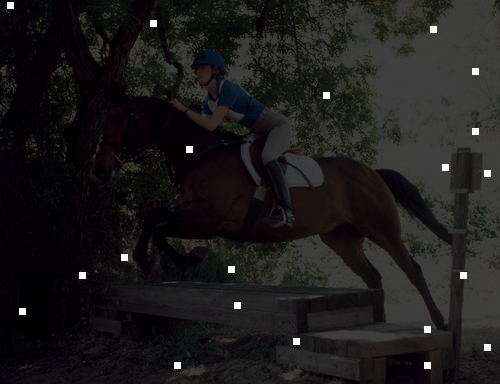}{person}{horse}
  \end{tabular}
  \caption{Examples of diverse sampling outputs. For each foreground class, we
    show the localization score map from global softmax model, and the
  selected 20 points. For background, the map shows the max over dot
  products with any selected foreground points.}
  \label{fig:sampling}
\end{figure*}

\begin{table*}[th]
  \centering
  \begin{tabular}{|l|l|l|l|}
    \hline

Method & VOC 2012 \val & VOC 2012 \test & comments\\
\hline
\textcolor{black}{DeepLab-CRF\cite{chen2015semantic}} & \textcolor{black}{68.7} &
                                                     \textcolor{black}{71.6}
                                        & \textcolor{black}{fully supervised}\\
\textcolor{black}{FCRN\cite{wu2016high}} & \textcolor{black}{74.8} &
                                                     \textcolor{black}{77.3}
                                        & \textcolor{black}{fully supervised}\\
\textcolor{black}{BoxSup\cite{dai2015boxsup}} & \textcolor{black}{62.0} &
                                                     \textcolor{black}{64.6}
                                        & \textcolor{black}{bounding box-supervised}\\

\textcolor{black}{Bbox-Seg\cite{papandreou2015weakly}} & \textcolor{black}{60.6} &
                                                     \textcolor{black}{62.2}
                                        & \textcolor{black}{bounding box-supervised}\\
\textcolor{black}{1 point~\cite{russakovsky2015what}} &
\textcolor{black}{35.1} &
                                                     \textcolor{black}{}&
\textcolor{black}{manual annotation, 1 pt/class}\\
\textcolor{black}{1 point+Obj~\cite{russakovsky2015what}}&
\textcolor{black}{42.7} &
                                                     \textcolor{black}{}&
\textcolor{black}{+ objectness prior}\\
\textcolor{black}{STC\cite{wei2015stc}} & \textcolor{black}{49.8} &\textcolor{black}{51.2} & \textcolor{black}{externally trained saliency model and 40K extra images}\\
\hline
MIL-sppx~\cite{pinheiro2015image} & 36.6 & 35.8 & superpixel smoothing\\
CCCN~\cite{pathak2015constrained} & 35.3 & 35.6 &\\
EM-Adapt~\cite{papandreou2015weakly} & 38.2 & 39.6 &\\
\hline
Ours & 40.6& 41.2& no post-processing\\
Ours & 45.2 & 46  & CRF post-processing, less than 3 minutes training time \\
\hline
SEC\cite{kolesnikov2016seed}& 50.7 & 51.5 & 7-8 hours training time\\
    \hline
  \end{tabular}
  \caption{Comparison of competitive segmentation methods, supervised with
    image-level tags. For reference, top of the table includes
    representative numbers methods trained
  with stronger supervision regimes on VOC 2012 data, or on additional
  data. 
}
  \label{tab:comparison}
\end{table*}

\section{Experiments}\label{sec:exp}
In order to compare our method to other work on segmentation, we
conduct all of our experiments on the VOC 2012 data set. For training
images (10,582 images in the \aug set) we discard all annotations except for image-level labels indicating which of
the 20 foreground classes is present in each image. We evaluate
various versions of our method, as well as its components
individually, on the \val set, and finally use the models chosen based
on tese experiments to obtain results on the \test set from the VOC
evaluation server. All experiments were done in Torch7, using Adam~\cite{kingma2015adam}
update rule when training networks.

\setlength{\parskip}{0cm}

\subsection{Experimental setup}

\paragraph{Pixel features}
As the base CNN we use the VGG-16 network trained on ImageNet and
publicly available from the authors of~\cite{simonyan2014very}, from
which we remove all layers above {\texttt pool5}.
This network is run in the fully convolutional mode on input images
resized to 256$\times$336 pixels. Then each of the 13 feature maps
(outputs of all convolutional layers, with pooling applied when
available) is resized to 1/4 of the input resolution, and concatenated
along feature dimensions. This produces a tensor in which each
location on the coarse $64\times84$ grid has a 4,224-dimensional
feature vector. This closely follows the hypercolumn extraction
protocol in~\cite{hariharan2015hypercolumns} (but using all layers)
and~\cite{mostajabi2015feedforward}, but without superpixel pooling.

When computing dot products in diversity
sampling~\eqref{eq:div-fg},\eqref{eq:div-bg} we normalize zoomout feature vectors
to unit norm in two stages: each feature dimension is normalized to be
zero-mean, unit variance over the entire
training set, then each feature vector is scaled to be unit Euclidean norm.

\paragraph{Localization models} For each class, the fully
convolutional localization
network consists of a 1$\times$1 convolutional layer with 1024 units,
followed by ReLU and the 1$\times$1 convolutional layer with 2 units,
which outputs the score maps $S$ for the foreground and $\bar{S}$ for background. At
training time, for the global softmax model~\eqref{eq:globalsoftmax}
this is futher followed by global max pooling layer and the softmax
layer, while for the per-pixel softmax~\eqref{eq:softmax} the order of
softmax and max pooling is reversed. 

For each class, we train the network on all positive examples for that
class (images that contain it) and a randomly sampled equal
number of negative examples, with
batch size of 1 image, learning rate $1e^{-4}$ which after 2 epochs is
\emph{decreased} to $1e^{-5}$ for one additional epoch
and momentum 0.9.

We experimented with adding higher layer features (\texttt{fc7} from
VGG-16) to the input to localization networks, but found that it makes
localization worse: it is too easy for the network to determine
presence of objects from these complex, translation invariant
features. We do however bring these features back when training the
final segmentation model, described next.

\paragraph{Segmentation model}
To provide image-level priors, which have been reported to improve
segmentation results in both fully supervised~\cite{mostajabi2015feedforward} and weakly
supervised~\cite{pinheiro2015image} settings, we augment the zoomout feature map with the 
global features (layer \texttt{fc7} of VGG-16, pooled over the entire
image and replicated for all locations). The combined
feature map (8320 features per location) is fed to a $1\times1$
conv. layer with 512 units, followed by ReLU, and the 21-channel
prediction layer, followed by the softmax loss layer. We train this
network on the selected set of points pooled over all training images, using batch size 100 (note:
this means 100 sample points, not 100 images!), fixed learning rate
$1e^{-6}$, and momentum 0.9, for two epochs. With these settings,
typical time to train the final segmentation model is less than 3 minutes
on a single Titan X GPU.

\subsection{Evaluation of model components}
We start by evaluating components of 
for our approach on the \val set.

\paragraph{Localization model} As shown in Table~\ref{tab:locmodels},
the global softmax model~\eqref{eq:globalsoftmax} obtains
significantly better results than the per-pixel
softmax~\eqref{eq:softmax}. Therefore we choose it for all the
subsequent experiments.

\paragraph{Self-supervision by localization maps} We could attempt using the score maps obtained by localization
network directly as the predicted segmentation maps. Specifically, we
considered assigning each pixel to the highest scoring class (after
normalizing the scores so that for each class on average images with
the foreground present have the highest score of 1), or to the
background if the highest foreground score is below
threshold. This results in poor segmentation accuracy: the highest
\val mIoU after sweeping the
threshold values was 25.66, for threshold of 0.2. 

We also made an attempt to use
these score-based segmentation maps as the source of self-supervision directly,
without sampling. That is, we can train the segmentation network in
the usual fully-supervised way, giving it the score map-based
dense segmentation labels as if it were ground truth. The results were
very poor; in large fraction of the pixels the localization models
uncertain, and while our sampling focuses
on high-score points, dense self-supervision is forced to make a
decision in those uncertain points as well, leading to a very noisy labeling.

\paragraph{Effect of sampling strategy}
For our diverse sampling method, we need to set the value of
$k$. Table~\ref{tab:miou-vs-k} shows the effect this value has on
\val accuracy. The optimal $k$ among those tested is 20, but the
behavior is stable across a large range of values. 

We also compared alternative sampling strategies, namely selecting the
top $k$ scoring points for each class, or using diversity but in
spatial domain instead of in feature domain. Table~\ref{tab:sampling}
shows that the diversity sampling using feature similarity is indeed superior
to those. Figure~\ref{fig:sampling} shows a few qualitative examples
of diverse sampling outputs. Notably, sampling for background is
usually quite accurate, even though it is oblivious to the actual
class scores and is entirely driven by diversity w.r.t. to the
foreground and within background.




\setlength{\tabcolsep}{4pt}

\newcommand{\setupSegEx}[1]{\includegraphics[width=.18\textwidth]{Figures/SegIms/#1} &%
\includegraphics[width=.18\textwidth]{Figures/SegIms/#1_GT}&%
\includegraphics[width=.18\textwidth]{Figures/SegIms/#1_base_heatmap}&%
\includegraphics[width=.18\textwidth]{Figures/SegIms/#1_our_model}&%
\includegraphics[width=.18\textwidth]{Figures/SegIms/#1_our_model_crf}\\}

\begin{figure*}[!th]
  \centering
  \begin{tabular}{ccccc}
    input & GT & loc-map & diverse 20 & diverse 20 with CRF\\
\setupSegEx{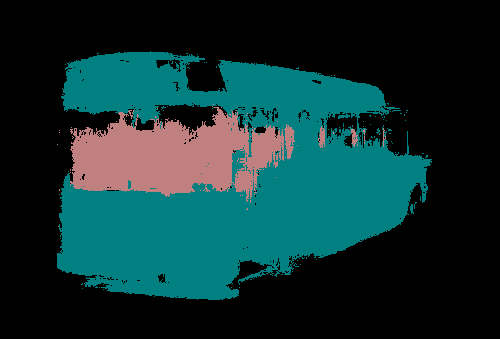}
    \setupSegEx{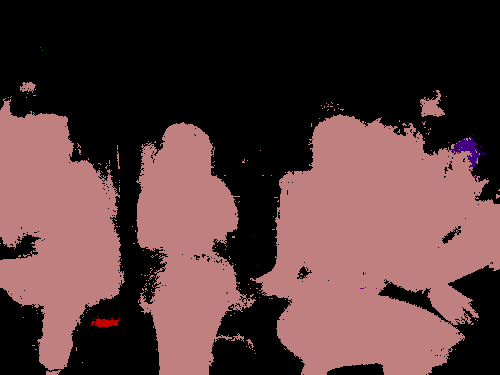}
    \setupSegEx{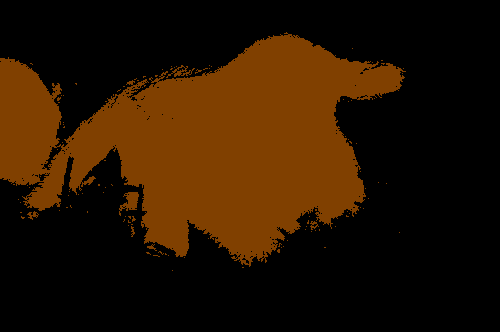}
    \setupSegEx{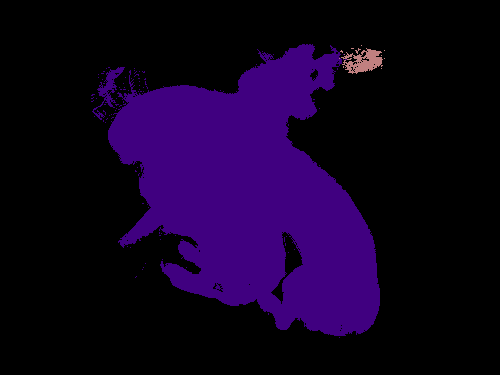}
\setupSegEx{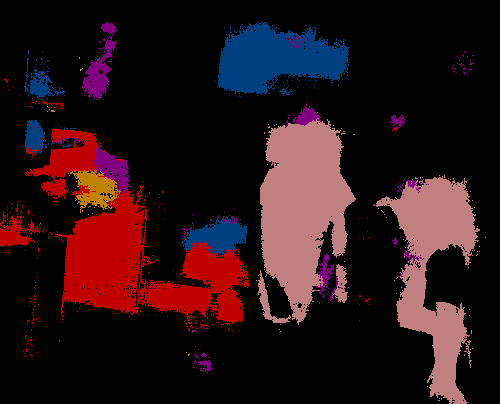}
\setupSegEx{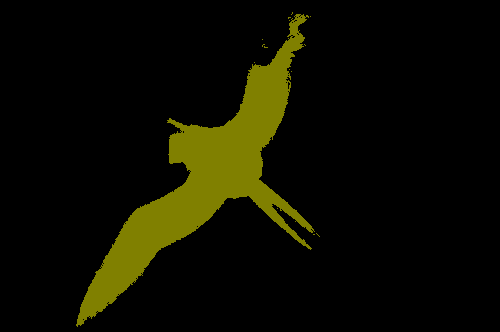}
\setupSegEx{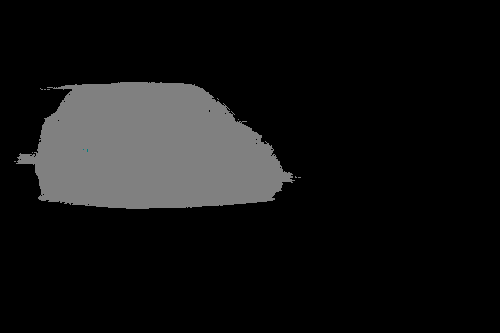}
\setupSegEx{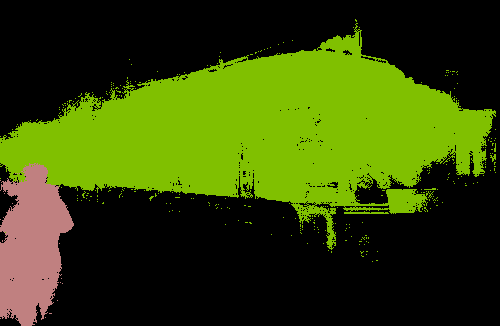}
  \end{tabular}
  \caption{Examples of segmentations learned through our
    self-supervision approach. From left: input image, ground truth,
    thresholded localization score maps, segmentation learned with diverse $k=20$ sampling, CRF post-processing.}
  \label{fig:examples}
\end{figure*}

\subsection{Final segmentation results}

Based on the preliminary experiments on \val above, we could identify
the optimal setting for our approach: global softmax localization model, with
diverse sampling, $k=20$. We report the results of
this method in Table~\ref{tab:comparison}. We also report the results of applying fully-connected CRF ~\cite{koltun2011efficient} with author's default parameters on top of our predictions.



The top portion of the table contains representative results for
stronger supervision scenarios, for reference; these are not directly
comparable to our results. Among the methods trained on the same data and in
the same regime as ours, our results are the highest. It is
interesting that we obtain similar results to those with
\emph{manual} annotation of a single point per class per image~\cite{russakovsky2015what} (better than their results without
objectness prior),
although our point selection is fully automatic.

We are aware of
additional results from concurrent work appearing on arXiv. The
STC~\cite{wei2015stc}, reporting \test mIoU of 51.2; it is trained
on 40,000 additional images, collected in a carefully designed
procedure to make them easy to learn from. STC also uses an externally
trained saliency mechanism, which requires mask annotations to
train. 

The only other method trained solely on VOC data which has higher accuracy is SEC~\cite{kolesnikov2016seed}, achieving 51.5 mIoU on \test. However, their approach is considerably more complex than ours,
employing hyperparameters that determine various thresholds, and the
time to train the final segmentation system with SEC is almost two
orders of magnitude higher than ours (7-8 hours vs. 3 minutes for
us). SEC also uses significantly larger field of view for the
underlying segmentation network than in our experiments (378 vs. 224
for us), and results reported in~\cite{kolesnikov2016seed} suggest
that this may be very important. We plan to investigate whether
increasing field of view of the segmentation network improves our
results as well.

Not only does our method obtain competitive results, but due to its very fast training of segmentation model, it is practical to add new classes on the fly, unlike other approaches.

We show some qualitative results obtained by our model on \val images
in Figure~\ref{fig:examples}.

\paragraph{Adding new object classes on the fly} 
One of the key characteristics of our model is modularity. Consider that we want to add new classes like Giraffe and Elephant which are not part of VOC dataset. We train localization model for new classes with only image level tags from MS-COCO dataset for Giraffe and Elephant. Since we have trained the localization model for each class separately, there is no need to re-train the other classes localization models. The segmentation training data for new classes is the sparse set of diverse points extracted from localization model output. Sparsity significantly speedup the segmentation training and diversity leads to high quality segmentation output. It takes less than 3 minutes to re-train the segmentation model with additional classes without hurting its accuracy on Pascal segmentation benchmark. Hence it is practical to add new classes on the fly. Qualitative examples of segmentation results for the new classes are shown in Figure~\ref{fig:extraseg}.

%% file: conclusions.tex
\section{Conclusions}\label{sec:conclusions}

We have proposed an approach to learning category-level semantic
segmentation when the only annotation available is tags indicating the
presence (or absence) of each foreground class in each image. Our approach is based on a form of
self-supervision, in which a sparse, visually diverse set of points in
training images is
labeled based on class-specific localization maps, predicted from the
image tags. 

Among the appealing properties of our method are its simplicity,
near absence of hyperparameters (and insensitivity to the only
hyperparameter, the self-supervision sample size), modularity (easy to
update the model with new classes and/or examples), lack of reliance
on complex external components requiring strong supervision, and last
but not least, competitive empirical performance and speed.

%% file: Acknowledgments.tex
\section*{Acknowledgments}\label{sec:Acknowledgments} 

We gratefully acknowledge a gift of GPUs from
NVIDIA corporation. GS was partially supported by NSF
award 1409837.